\documentclass[preprint,12pt]{elsarticle}

\usepackage{amssymb}
\usepackage{amsmath} 

\journal{a journal}

\renewcommand{\tilde}{\widetilde} 

\begin{document}

\begin{frontmatter}



\title{Hidden Unit Specialization in Layered Neural Networks: ReLU vs. Sigmoidal Activation}


\author{Elisa Oostwal, Michiel Straat, Michael Biehl}

\address{Bernoulli Institute for Mathematics, Computer Science
and Artificial Intelligence\\
University of Groningen, Nijenborgh 9, 9747 AG Groningen, The Netherlands\\
e-mail: {\rm e.c.oostwal@student.rug.nl, m.j.c.straat@rug.nl, m.biehl@rug.nl} }

\begin{abstract}
By applying concepts from the statistical physics of learning, we 
study layered neural networks of rectified linear 
units (ReLU). The comparison with conventional, sigmoidal activation 
functions is in the center of interest. 
We compute typical learning curves for large shallow networks 
with  $K$ hidden units in matching student teacher scenarios. 
The systems undergo phase transitions, i.e.\ sudden changes of the generalization 
performance via the process of hidden unit specialization 
at critical sizes of the training set. Surprisingly, our results show that the 
training behavior of  ReLU networks  is qualitatively different from that of networks with 
sigmoidal activations.  In networks with $K\geq 3$  sigmoidal hidden units, 
the transition is discontinuous: Specialized network configurations co-exist 
and compete with states of poor performance even for very large training sets. 
On the contrary, the use of ReLU  activations results in continuous transitions 
for all $K.$ For large enough training sets, two competing, differently specialized 
states display similar generalization abilities, which coincide exactly for large 
hidden layers in the limit $K\to\infty.$  Our findings are also confirmed in 
Monte Carlo simulations of the training processes. 
\end{abstract}



\begin{keyword}
Neural Networks, Machine Learning, Statistical Physics
\end{keyword}
\end{frontmatter}


\section{Introduction}
\label{introduction}
The re-gained
interest in artificial neural networks
\citep{hertz, bishop1995, EngelvandenBroeck, hastie, bishop2006}
is largely due to the successful application 
of so-called Deep Learning 
in a number of practical contexts, see e.g.\ 
\cite{goodfellow,naturedeep,esanndeep} for reviews and
further references.

The successful training of powerful, multi-layered {deep} networks
has become feasible for a number of reasons including the
automated acquisition of large amounts of training data in various domains, 
the use of modified and optimized architectures,
e.g.\ convolutional networks for image processing, 
and the ever-increasing availability of computational power 
needed for the implementation of efficient training. 

One particularly important modification of  earlier
models is the use of alternative activation functions
\cite{goodfellow,searching,timetoswish}. 
Arguably, so-called rectified linear units (ReLU) constitute 
the most popular choice in Deep Neural Networks 
\cite{goodfellow,searching,timetoswish,acousticmodels,hahnloser,krizhevsky}. 
Compared to more traditional activation functions,
the simple ReLU and recently
suggested modifications warrant computational ease and appear to
speed up the training, see for
instance \cite{acousticmodels,nair,villmannswish}. 
The one-sided ReLU
function is found to yield sparse activity in large networks,
a feature which is frequently perceived as favorable and biologically
plausible \cite{goodfellow,hahnloser,sparse}.
In addition, the problem of 
\textit{vanishing gradients}, which arises 
when applying the chain rule
in layered networks of sigmoidal units, 
is avoided \cite{goodfellow}.  
Moreover, networks of 
rectified linear units have displayed
favorable generalization behavior in several practical
applications and benchmark tests,
e.g.\  \cite{searching,timetoswish,acousticmodels,hahnloser,krizhevsky}. 

The aim of this work is to 
contribute 
to a better theoretical
understanding of how the use of
ReLU activations influences and
potentially improves the training behavior 
of layered neural networks. We focus on
the comparison with 
traditional 
sigmoidal functions and
analyse non-trivial model situations.  
To this end,  we employ
approaches from the statistical physics of learning,
which have been applied earlier with
great success
in the context of neural networks and machine learning
in general \cite{hertz,EngelvandenBroeck,SST,revmodphys,kinzel,opper,handbook}.
The statistical physics approach complements other theoretical
frameworks in that it studies the typical behavior of large 
learning systems in model scenarios. As an important example, learning
curves have been computed in a variety of settings, 
including on-line and off-line supervised training of 
feedforward neural networks, see
for instance 
\cite{EngelvandenBroeck,SST,revmodphys,kinzel,opper,handbook,schwarzegradient,saadsollashort,saadsollalong,woehlertransient,nestorplateaus,retarded} and references therein.
A topic of particular interest for this work
is the analysis of phase transitions
in learning processes, i.e.\  sudden changes
of the expected performance with the 
training set size or other 
control parameters, see  \cite{kinzel,opper,retarded,Kang,dpg,epl,epj,phasetransitions}
for examples and further references. 

Currently, the statistical physics of
learning is being revisited extensively
in order to investigate relevant phenomena in deep neural 
networks and other learning paradigms, see 
\cite{monasson,kadmon,pankaj,sohl,egalitarian,esannsession2019,relucapacity,goldt} 
for recent examples and further references. 

In this work, we systematically study
the training of layered networks in so-called student teacher 
settings, see e.g.\ \cite{EngelvandenBroeck,SST,revmodphys,opper}. We consider idealized, yet non-trivial 
scenarios of matching student and teacher complexity. 
Our findings demonstrate
that ReLU networks display training and
generalization behavior which differs significantly from their
counterparts composed of 
sigmoidal units. Both network types display
sudden changes of their performance 
with the number of available examples. 
In statistical physics terminology,
the systems undergo phase transitions at a 
critical training set size.
The underlying process of hidden unit 
specialization 
and the existence of saddle points in the objective function 
have recently attracted attention also in the context of
Deep Learning \cite{kadmon,dauphin,saxe}.

Before analysing ReLU networks, 
we confirm  earlier theoretical results which 
indicate that the 
transition for large networks of sigmoidal units 
is discontinuous 
\textit{(first order)}:
For small training sets, 
a poorly generalizing state is
observed, in which all hidden units approximate
the target to some extent and
essentially perform the same task. 
At a
critical size of the training set,
a favorable configuration with 
specialized
hidden units appears. However, 
a poorly performing state 
remains metastable  and the 
specialization
required for successful learning can delay the training process significantly 
\cite{Kang,dpg,epl,epj}. 

In contrast we find that, surprisingly,  
the corresponding phase transition in 
ReLU networks is always continuous \textit{(second 
order)}. At the transition, the unspecialized 
state is replaced by two competing 
configurations with very similar generalization
ability. In large networks, their performance is
nearly identical and it coincides exactly in the
limit $K\to\infty.$

In the next section we detail the 
considered models and outline the theoretical
approach. In Sec.\ \ref{results} our 
results are presented and discussed. In addition,
results of supporting Monte Carlo simulations are presented. 
We conclude with a summary and outlook on future extensions
of this work.

\section{Model and Analysis} \label{methods}
Here we introduce the modelling framework, i.e.\
the considered student teacher scenarios. Moreover,
we outline their analysis by means of statistical 
physics methods and discuss the simplifying
assumption of training at high (formal) temperatures.

\subsection{Network architecture and
activation functions}
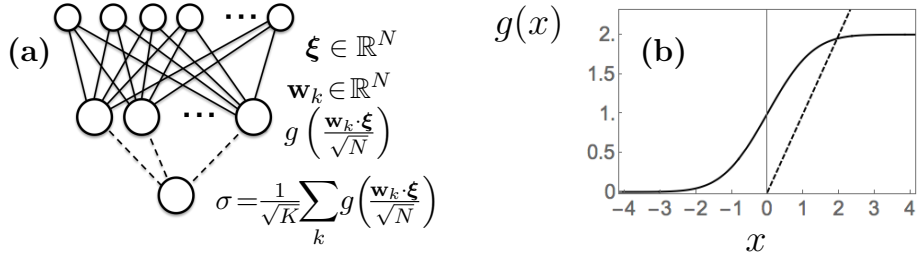
\begin{figure}[t]
\begin{center} 
  \includegraphics[width= 0.24\textwidth]{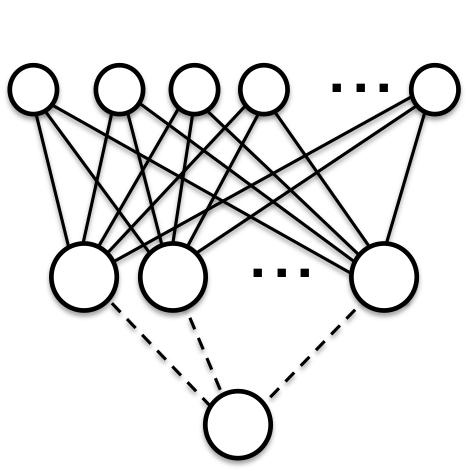} 
 \put(-110,60){{\bf (a)}} 
  \put(3,62){\mbox{\small $\boldsymbol{\xi}\in \mathbb{R}^N$}}
  \put(-8,46){
  \mbox{\small $\mathbf{w}_k\!\in\! \mathbb{R}^N 
  $}}
  \put(-5,30){\mbox{\small $g\left(\frac{\mathbf{w}_k\cdot\boldsymbol{\xi}}{\sqrt{N}}\right)$}}
  \put(-31,4){\mbox{\small $\textstyle \sigma \!=\!\!
  \frac{1}{\sqrt{K}}\!\!  \displaystyle 
     \sum_{k}\!
     \textstyle g\!\left(\!\frac{\mathbf{w}_k\cdot\boldsymbol{\xi}}{\sqrt{N}}\right) $}}
\mbox{~~~~~~~~~~~~~~~~~~~~~~~~~~~}    
 \includegraphics[width= 0.32\textwidth] {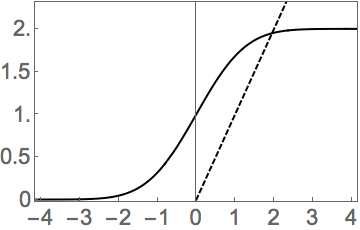} 
\put(-105,60){{\bf (b)}} 
  \put(-160,70){\mbox{\large$g(x)$}}
  \put(-65,-12){\mbox{\large $x$}} 
 \end{center} 
  \caption{{\bf (a)} Illustration of the  
  network  architecture with an $N$-dim.
  input layer, a set of adaptive weight vectors
   $\mathbf{w}_{k} \mbox{~with~} k\!=\!1,\!\ldots\!,K$ 
  (represented by solid lines)
  and total output $\sigma$ given by the sum
  of hidden unit activations with 
  fixed weights (dashed lines). 
  {\bf (b)} The
  considered activation functions: the 
  sigmoidal $g(x)=\left(1+\mbox{erf}[x/\sqrt{2}]\right)$
  (solid line) 
  and the ReLU activation $g(x)=\mbox{\rm max}\{0,x\}$ (dashed line).
   \label{FIG1}}
  \end{figure}

We consider feed-forward neural networks 
where $N$  input nodes represent feature vectors
$\boldsymbol{\xi}\in \mathbb{R}^N.$ A single
layer of $K$ hidden units is connected
to the input through adaptive weights
$\underline{W}=\left\{\mathbf{w}_k \in \mathbb{R}^N
\right\}_{k=1}^K.$ The total real-valued output reads
\begin{equation} \label{studentoutput}\sigma(\boldsymbol{\xi}) 
 = \frac{1}{\sqrt{K}} \sum_{k=1}^K \, 
   g\left(x_k\right) \mbox{~~with~~}
x_k =    \frac{1}{\sqrt{N}} \mathbf{w}_k \cdot \boldsymbol{\xi}.
 \end{equation} 
The quantity $x_k$ is
referred to as the
\textit{local potential} of
the hidden unit. 
The resulting 
activation is specified
by the function $g(x)$ and hidden
to output weights are fixed
to $1/\sqrt{K}$. 
Figure \ref{FIG1} (a) 
illustrates the network architecture.
  
This type of network has been termed
the \textit{Soft Committee Machine\/} (SCM)
in the literature due to its vague  
similarity to the committee machine for
binary classification, e.g.\  \cite{EngelvandenBroeck,revmodphys,opper,retarded,relucapacity,robertcapacity,holm}. There, the discrete output
is determined by the majority of threshold
units in the hidden layer, while the SCM
is suitable for regression tasks.

We will consider two popular types of
transfer functions:
\begin{itemize}
\item[a)] {\bf Sigmoidal activation}\\
Frequently, S-shaped transfer functions 
$g(x)$ have
been employed,  
which increase  
monotonically  
from \textit{zero} at large
negative arguments 
and saturate at a finite maximum
for $x\to\infty$. Popular examples are based on $\tanh(x)$ or 
the sigmoid $(1+e^{-x})^{-1}$, 
often with an additional threshold $\theta$  
as in $g(x-\theta),$  or a
steepness parameter controlling the magnitude
of the derivative $g^\prime.$ 
We study in particular
\begin{equation} \label{erfactivation} \textstyle 
g(x) = \left( 1 + \mbox{erf}\left[
\frac{x}{\sqrt{2}}\right] \right) 
= 2 \int\limits_{-\infty}^{x} dz \frac{e^{-z^2/2}}{\sqrt{2\pi}} \end{equation}
with $0\leq g(x) \leq 2,$ 	
which is displayed in Fig.\ \ref{FIG1} (b). 
The relation to an
integrated Gaussian facilitates
significant mathematical ease, which has been
exploited in numerous studies 
of machine learning models, e.g.\ 
\cite{schwarzegradient,saadsollashort,saadsollalong}. Here, 
the function (\ref{erfactivation}) serves
as a generic example of a 
sigmoidal and 
its specific form
is not expected to influence our
findings crucially. 
As we argue below, the choice of limiting
values $0$ and $2$ for small and large arguments,
respectively, is also arbitrary and
irrelevant for the qualitative results of our
analyses. 
 \\

\item[b)] {\bf Rectified Linear Unit (ReLU) activation} \\
This particularly simple, piece-wise linear
transfer function has attracted considerable
attention in the  
context of multi-layered  neural networks:  
\begin{equation} \label{reluactivation}
g(x) = \, \mbox{max}\left\{ 0, x \right\} 
= \left\{ 
\begin{array}{cc} 0 & \mbox{~~for~} x\leq 0\\
                  x & \mbox{~~for~} x > 0 \\ 	
       \end{array} \right. 
\end{equation}
which is illustrated in Fig.\ \ref{FIG1}
(b). 
In contrast to sigmoidal activations, the 
response of the unit
is unbounded for $x\to\infty$.

The function
(\ref{reluactivation}) is obviously
not differentiable in $x=0$. Here, we can 
ignore this mathematical subtlety 
and remark that it is considered 
irrelevant in practice \cite{goodfellow}.
Note also that our theoretical
investigation in Sec.\ \ref{methods} does
not relate to a particular realization 
of gradient-based training. 

\end{itemize}
\noindent 
It is important to realize that 
replacing the above functions
by $g(x)= \gamma \, 
\left(1+\mbox{erf}[x/\sqrt{2}]\right)$ in (a) or
by $g(x)= \max\{0, \gamma \, x \} = \gamma \, \max\{0,x\} $
in (b), where $\gamma>0$ is an arbitrary factor, 
would be equivalent to
setting the hidden unit weights to 
$\gamma/\sqrt{K}$ in Eq.\ (\ref{studentoutput}). 
Alternatively, we could incorporate the factor $\gamma$ in
the effective temperature parameter
$\alpha$ of the theoretical analysis
in Sec.\ \ref{hightempsection}. 
Apart from this trivial re-scaling, 
our results would not be affected
qualitatively.

\subsection{Student and teacher scenario}
We investigate the training and
generalization behavior of the
layered networks introduced above 
in a setup that models the learning
of a regression scheme from example data. 
Assume that a given training set 
\begin{equation} \label{dataset} 	\textstyle
\mathbb{D} = \left\{ \boldsymbol{\xi}^\mu, \tau^\mu 
 \right\}_{\mu=1}^P  
\end{equation}
comprises $P$ input output pairs
which reflect the target task. In order to
facilitate successful learning, $P$
should be proportional to the
number of adaptive weights in the trained system. 
In our specific model scenario the labels
$\tau^\mu= \tau(\boldsymbol{\xi}^\mu)$ are
thought to be provided by a teacher SCM,  
representing the target 
input output relation 
 \begin{equation} \label{teacheroutput} \tau(\boldsymbol{\xi})  \textstyle 
 = \frac{1}{\sqrt{M}} \sum_{m=1}^M \, 
   g\left(x_m^*\right) \mbox{~~with~~}
   x_m^* = \frac{1}{\sqrt{N}} \mathbf{w}_m^* \cdot \boldsymbol{\xi}.
  \end{equation} 
The response
is specified in terms of the 
set of teacher
weight vectors $\underline{W}^* = \left\{ \mathbf{w}_m^* \right\}_{m=1}^M$ and defines the correct target
output for every possible feature vector
$\boldsymbol{\xi}$. 
For simplicity,  we  focus on
settings with orthonormal teacher weight
vectors and restrict the 
adaptive
student configuration to normalized weights: 
 \begin{equation}  \label{normalization} 
\mathbf{w}_m^* \cdot \mathbf{w}_n^* 
/N = \delta_{mn} \mbox{~~and~~} 
|\mathbf{w}_j|^2=N
\mbox{~~with the Kronecker-Delta~}
\delta_{mn}. 
 \end{equation}

Throughout the following, the
evaluation of the student network
will be based on a simple 
quadratic error measure that compares
student output and target value.
Accordingly, the selection of student
weights $\underline{W}$ in the training process
is guided by a cost  function which
is given by the corresponding  
sum over all available data in $\mathbb{D}$:  
\begin{equation} \label{costfunction}	
 E = \sum_{\mu=1}^P \epsilon (\boldsymbol{\xi}^\mu) 
\mbox{~~with~~} \epsilon(\boldsymbol{\xi})
 = \frac{1}{2} \bigg(\sigma(\boldsymbol{\xi}) - \tau(\boldsymbol{\xi}) \bigg)^2. 
\end{equation} 
By choosing
the parameters $K$ and $M$, a variety
of situations can be modelled. 
This includes
the learning of unrealizable rules ($K<M$) and
training of over-sophisticated
students with $K>M$.  
Here, we restrict ourselves to the idealized,
yet non-trivial case of perfectly matching
student and teacher complexity, i.e.\ $K=M$, which
makes it possible to achieve $\epsilon(\boldsymbol{\xi})=0$
for all input vectors. 

\subsection{Generalization error and order parameters} 
Throughout the following we  consider
feature vectors $\boldsymbol{\xi}^\mu$
in the training set with uncorrelated
i.i.d. random components of zero mean
and unit variance. Likewise, arbitrary input vectors 
 $\boldsymbol{\xi} \not\in
 \mathbb{D}$ 
 are assumed to follow the same statistics: 
$$ \left\langle\xi_j^\mu \right\rangle = 0, \,  \left\langle \xi_j^\mu \xi_k^\nu
\right\rangle = \delta_{j,k} \delta_{\mu,\nu},   \, 
 \left\langle\xi_j \right\rangle = 0 
\mbox{~and~} \left\langle \xi_j \xi_k
\right\rangle = \delta_{j,k}. $$ 
As a consequence of this assumption, the 
Central Limit Theorem applies to  
the local potentials 
 $x_j^{(*)} =\mathbf{w}_j^{(*)} \cdot 
 \boldsymbol{\xi}/\sqrt{N}$ which become correlated Gaussian 
 random variables of order ${\cal O}(1)$.  It is straightforward to
work out the averages $\left\langle \ldots \right\rangle$ and
(co-)variances:
\begin{equation} \label{covs} 
\left\langle x_k \right\rangle
= \left\langle x_k^* \right\rangle = 0,
\mbox{~} \left\langle x_j x_k \right\rangle = \, 
  \mathbf{w}_j \cdot \mathbf{w}_k / N 
 \, \equiv \, Q_{jk}, 
\end{equation} 
$$
 \left\langle x_j^* x_k^* \right\rangle = 
 \mathbf{w}^*_j \cdot \mathbf{w}_k^* / N 
  =  \delta_{jk} 
 \mbox{~and~} 
 \left\langle x_j x_k^* \right\rangle  = 
 \mathbf{w}_j \cdot \mathbf{w}_k^* / N 
 \equiv \, R_{jk},
 $$ 
 which fully specify the joint density
 $P\left( \left\{ x_i, x_i^* \right\} \right)$. 
 The \textit{order parameters}  
 $R_{ij}$ and $Q_{ij}$ for
 $(i,j=1,2,\ldots K)$ serve 
 as macroscopic characteristics of the student
 configuration. 
The norms $Q_{ii}=1$ are fixed according
to Eq.\ (\ref{normalization}), while  the symmetric 
 $Q_{ij}=Q_{ji}$ quantify the $K(K-1)/2$ pairwise alignments of 
 student weight vectors. 
 The similarity of the student weights to
 their counterparts
 in the teacher network are measured in terms
 of the $K^2$ quantities $R_{ij}$. Due to
 the assumed normalizations, the relations
 $-\!1\!\leq\!Q_{ij}, R_{ij}\!\leq\!1$ 
 are obviously satisfied.

Now we can work out the generalization error,
i.e.\ the expected deviation
of student and teacher output for a random
input vector, given specific
weight configurations $\underline{W}$ and
$\underline{W}^*$. 
Note that
SCM 
with $g(x)=\mbox{erf}[x/\sqrt{2}]$ 
have been treated in \cite{saadsollashort,saadsollalong}
for general $K,M.$  
Here, we resort to the special case of matching
network sizes, $K=M,$  with  
\begin{eqnarray} \label{epsggeneral} 
 \epsilon_g \, &=& \,
\frac{1}{2K} \left\langle 
\bigg(  \sum_{i=1}^K g(x_i) 
 - \sum_{j=1}^K g(x_j^*)\bigg)^2 \right\rangle. 
\end{eqnarray} 
We note here that matching additive constants
in the student and teacher activations would 
leave $\epsilon_g$ unaltered. 
As detailed in the Appendix, all averages in Eq.\ (\ref{epsggeneral}) 
can be computed analytically for both choices
of the activation function $g(x)$ in student
and teacher network. 
Eventually, the generalization error is
expressed in terms of very few macroscopic
order parameters, instead of explicitly 
taking into account $K N$ individual 
weights. 
The concept is characteristic for the statistical physics approach
to systems with many degrees of freedom. 
In the following, 
 we restrict the  analysis to student configurations which are
site-symmetric with respect to the hidden units:
\begin{equation} \label{sitesymmetry}
R_{ij} =  \left\{ \begin{array} {ll}
 R & \mbox{for~} i=j \\	
 S & \mbox{otherwise,} \\ 
 \end{array} \right.
\mbox{~~~~}
Q_{ij} =  \left\{ \begin{array} {ll}
 1 & \mbox{for~} i=j \\	
 C & \mbox{otherwise.} \\ 
 \end{array} \right.
\end{equation}
Obviously, the system is invariant under 
permutations, so we can restrict ourselves to one specific
case with matching indices $i=j$
in Eq.\ (\ref{sitesymmetry}). 
While this assumption reflects the symmetries of the student teacher scenario,
it allows for the \textit{specialization} of hidden units:  
For $R=S$ all student units display
the same overlap with all teacher 
units. In specialized configurations 
with
$R\neq S$, however, each student weight 
vector has achieved a 
distinct overlap with exactly
one of the teacher units. Our analysis 
shows
that states with both positive ($R>S$) and
negative specialization ($R<S$) can
play a
significant role in the training process. 
Under the above assumption of site-symmetry 
(\ref{sitesymmetry}) and applying the 
normalization (\ref{normalization}), the 
generalization error (\ref{epsggeneral}),
see also Eqs.\ 
(\ref{egsigmostgeneral},\ref{egrelumostgeneral}),  becomes \\

\noindent 
a) for  
$ g(x)= \left(1\!+\!\mbox{erf}\big[{x/\sqrt{2}}\big]\right)$
in student and teacher \cite{saadsollashort}: 
\begin{equation} \label{egsigsiteerf} \textstyle 
 \epsilon_g \!=\! \frac{1}{K}  \left\{\frac{1}{3}\!+\!
\frac{K\!-\!1}{\pi}\left[ \sin^{-1}\left(\frac{C}{2}\right)\!-\! 2\sin^{-1}\left(\frac{S}{2}\right)\right]
 \!-\!\frac{2}{\pi}\sin^{-1}\left(\frac{R}{2}\right)\right\},
\end{equation}
\noindent 
b) for ReLU  $ g(x)= \mbox{max}\{0,x\}$ \cite{straat2019}: 
\begin{eqnarray} \label{egsigsiterelu} 
 \epsilon_g &=& \textstyle 
\frac{1}{2K} \left\{
K\!+\!\frac{(K^2\!-\!K)}{2\pi}\!-\!2K\!\left(\frac{R}{4}+\frac{\sqrt{1-R^2}}{2\pi} + \frac{R \sin^{-1}(R)}{2\pi} \right) 
\right.   \\ &+& \!\!\!\!
\textstyle (K^2\!-\!K)\left(\frac{C}{4}\!+\!\frac{\sqrt{1-C^2}}{2\pi}\!+\!
\frac{C \sin^{-1}(C)}{2\pi} \right)  
\nonumber 
- \textstyle \!\! \left. 2(K^2\!-\!K) \left(\frac{S}{4}\!+\!\frac{\sqrt{1-S^2}}{2\pi} \!+\! \frac{S \sin^{-1}(S)}{2\pi} \right) 
 \right\}.
 \label{egrelusite} 
\end{eqnarray}
In both settings,
perfect agreement of student
and teacher with $\epsilon_g=0$ is
achieved for  $C\!=\!S\!=\!0$ and $R\!=\!1.$ 
The scaling of 
outputs with hidden to output weights
$1/\sqrt{K}$  in Eq.\ (\ref{studentoutput})
results in a generalization error which is not explicitly $K$-dependent
for uncorrelated random students:
A configuration with $R=C=S=0$ yields
$\epsilon_g=1/3$ in the case of sigmoidal activations (a), whereas 
$\epsilon_g = \frac{1}{2}-\frac{1}{2\pi}\approx 0.341$ for  
ReLU student and teacher.

\subsection{Thermal equilibrium and the high-temperature limit} \label{hightempsection} 
In order to analyse the expected outcome
of training from a set of examples $\mathbb{D}$, 
we follow the well-established statistical 
physics approach and analyse an 
\textit{ensemble} of networks in a
formal \textit{thermal equilibrium} situation.
In this framework, the cost function $E$
is interpreted as the {\sl energy} of 
the system and the density of observed network states is given
by the so-called Gibbs-Boltzmann density 
\begin{equation}	\label{gibbs}  \textstyle 
\left.{\exp[-\beta E]}\right/{Z}  \mbox{~~~with~~}
 Z=\int d\mu(\underline{W}) \exp[-\beta E],
\end{equation}
 where the measure $d\mu(\underline{W})$ 
 incorporates potential restrictions of the 
 integration over all possible configurations 
 of 
 $\underline{W}=\left\{\mathbf{w}_i\right\}_{i=1}^K,$  
 for instance the normalization
 $\mathbf{w}_k^2 = N$ for all $k$.   
 This equilibrium density could result from 
 a Langevin type of training dynamics
 $$ \left.{\partial \underline{W}}\right/{\partial t}  \, = 
  \, - \nabla_W \, E(\underline{W}) \, + \underline{\eta}, $$
where $\nabla_W$ denotes the
gradient with respect to 
all $KN$ 
degrees of freedom in the student network. 
Here, the minimization of $E$ is performed in 
the presence  of a $\delta$-correlated, zero mean 
noise term $\underline\eta(t)\in \mathbb{R}^{KN}$ with
$$ \left\langle\eta_i(t)\right\rangle=0
\mbox{~~and~~} \left\langle\eta_i(t)\eta_j\right(t^\prime)\rangle = 2 \beta \delta_{ij} \delta(t-t^\prime),$$
where $\delta(\ldots)$ denotes the Dirac delta-function.
The parameter $\beta=1/T$ controls 
the strength of the \textit{thermal noise}
in the gradient-based minimization of $E$. 
According to the, by now, standard statistical physics approach to
off-line learning \cite{hertz,EngelvandenBroeck,revmodphys,SST}
typical properties of the system are 
governed by the so-called \textit{quenched free energy} 
\begin{equation}
    \label{fdef} 
f = -\left.{1}\right/{N} \,  \left\langle \, \ln Z \right\rangle_{\mathbb{D}} / \beta
\end{equation} 
where $\left\langle \ldots \right\rangle_{\mathbb{D}}$ denotes the average
over the random realization of the training set. 
In general, the evaluation of the
quenched average $\left\langle \ln Z \right\rangle_{\mathbb{D}}$ is technically
involved and requires, for instance, the application of the 
replica trick \cite{hertz,EngelvandenBroeck,revmodphys}.
Here, we resort to the simplifying limit of training
at high temperature $\ T\to \infty, \beta\to 0$, which has proven
useful in the qualitative investigation of various 
learning scenarios \cite{SST}. 
In the limit $\beta\to 0$ the so-called
annealed approximation \cite{SST,revmodphys,EngelvandenBroeck}
$\left\langle \ln Z \right\rangle_{\mathbb{D}} 
\approx \ln \left\langle Z \right\rangle_{\mathbb{D}}$ becomes exact.  
Moreover,  we have 
\begin{equation} 
\textstyle  \left\langle Z \right\rangle_{\mathbb{D}}  = 
\left\langle \int d\mu(\underline{W}) 
 e^{-\beta  E }   \right\rangle_{\mathbb{D}} \approx  
 \int d\mu(\underline{W}) e^{-\beta \left\langle E \right\rangle_{I\!\!
D}}.
\end{equation} 
Here, $P$ is the number of statistically   
independent examples in $\mathbb{D}$
and 
 $\left\langle E\right\rangle_{\mathbb{D}} = P \left\langle \epsilon(\boldsymbol{\xi}) \right\rangle_\xi = P \epsilon_g$. 
As the exponent grows
linearly with $P\propto N$, the integral is dominated 
by the maximum of the integrand. 
By means of a saddle-point 
integration for $N\to\infty$ we obtain  
\begin{equation} \label{ffirst} 
 -\frac{1}{N} \ln \left\langle Z\right\rangle_{\mathbb{D}} \, =  \beta f(\{R_{ij},Q_{ij}\})\approx  \frac{\beta P}{N} \epsilon_g  - s.
\end{equation} 
Here, the right hand side has to be minimized 
with respect to the arguments, i.e.\ the order parameters 
$\left\{R_{ij},Q_{ij}\right\}.$  
In Eq.\ (\ref{ffirst}) 
we have introduced the entropy term 
\begin{equation} \label{entropydef}  \textstyle 
 s\!=\!\frac{1}{N}\!\ln\!\int\!d\mu(\underline{W})\!
 \prod_{i,j}\!\left[\delta(NR_{ij}\!-\!\mathbf{w}_i\cdot\mathbf{w}^*_j)\delta(NQ_{ij}\!-\!\mathbf{w}_i\!\cdot\!\mathbf{w}_j)\right]. 
\end{equation} 
The quantity $e^Ns$
corresponds to the volume in weight space 
that is  consistent
with a given configuration of order parameters. 
Independent of the activation functions or
other details of the learning problem, one 
obtains for large $N$ \cite{epl,epj}
\begin{equation} \label{slogdet} 
s(\left\{R_{ij},Q_{ij}\right\}) =  \ln 
\det \left({\cal C}\right) /2 
+ \mbox{const.}
\end{equation} 
where ${\cal C}$ is the $(2K\times 2K)$-dimensional 
matrix of
all pair-wise and self-overlaps of the vectors $\left\{\mathbf{w}_i, \mathbf{w}_i^*\right\}_{i=1}^K$, i.e. the matrix of all 
$\left\{R_{ij}, Q_{ij}, T_{ij}\right\},$ see also Eq. 
(\ref{fullmatrix}) in the Appendix. 
The constant term is independent
of the order parameters and, hence, irrelevant for
the minimization in Eq.\ (\ref{ffirst}).  A compact 
derivation of (\ref{slogdet}) is provided in, e.g., \cite{epj}. 

Omitting additive constants and assuming the normalization
(\ref{normalization}) and site-symmetry (\ref{sitesymmetry}),
the entropy term reads \cite{epl,epj}
\begin{equation} \label{sK} \textstyle 
s=\frac{1}{2}  \ln \left[ 1\!+\!(K\!-\!1)C\!-\!\left((R\!-\!S)\!+\!K S\right)^2 \right] 
+ {\textstyle \frac{K\!-\!1}{2}} \displaystyle \ln \left[ 1\!-\!C\!-\!(R\!-\!S)^2  \right].
\end{equation}

In order to facilitate the successful adaptation of $KN$  weights
in the student network we have to assume that the number of examples also scales like  $ P = \widetilde{\alpha} \, K\, N. $  
Training at high temperature additionally requires that 
$\alpha = \widetilde{\alpha} \beta = {\cal O}(1)$
for $\widetilde{\alpha}\to\infty, \beta\to 0,$
which yields a free energy of the form 
\begin{equation} \label{betafRSC}
\beta \, f(R,S,C) \, = \, \alpha \, K \, \epsilon_g(R,S,C) \, - \, s(R,S,C).
\end{equation} 
The quantity $\alpha= \beta P/(KN)$ can be interpreted as
an effective temperature parameter or, likewise, as the properly 
scaled training set size. The high temperature has to be compensated
by a very large number of training examples in order to facilitate
non-trivial outcome. As a consequence, the energy of the system 
is proportional to $\epsilon_g$, which implies
that  training and generalization error
are effectively identical in
the simplifying limit. 

\begin{figure}[t!]
  \begin{center}	
\hfill 
  \includegraphics[width= 0.43\textwidth]{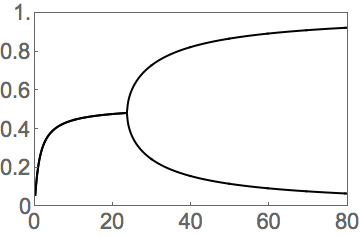} 
  \put(-190,95){$R,S$} 
 \put(-145,88){\bf (a)} 
  \put(-70,0){\large $\alpha$} 
  \put(-60,+58){$R\!\neq\!S$}
  \put(-140,+42){$R\!=\!S$}
\mbox{~~~~~~}
   \includegraphics[width= 0.43\textwidth]{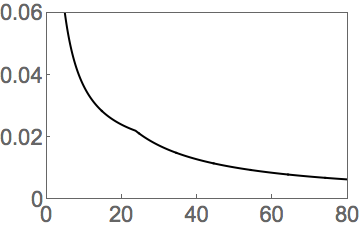}
 \put(-185,95){\large $\epsilon_g$} 
\put(-125,88){{\bf (b)}} 
 \put(-50,88){$K=2$}
  \put(-68,0){\large $\alpha$} 
  \ \\[5mm] 
\hfill 
  \includegraphics[width= 0.42\textwidth]{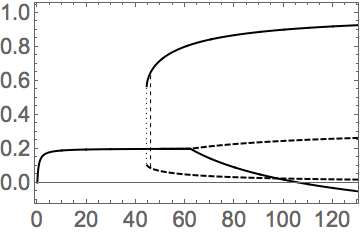}
\put(-135,88){{\bf (c)}} 
\put(-190,95){$R,S$} 
  \put(-55,-4){\large $\alpha$} 
  \put(-140,+43){$R\!=\!S$}
  \put(-40,+82){$R\!>\!S$}
  \put(-40,+48){$R\!<\!S$}
\mbox{~~~~~}
  \includegraphics[width= 0.44\textwidth]{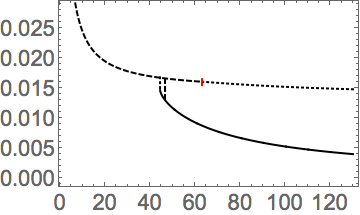} \mbox{$\,$}
\put(-130,88){{\bf (d)}} 
 \put(-185,100){\large $\epsilon_g$} 
  \put(-60,-4){\large $\alpha$} 
 \put(-50,88){$K=5$}
  \put(-40,35){$R\!>\!S$}
  \put(-110,+68){$R\!=\!S$}
  \put(-76,65){$R\!<\!S$}
 \end{center}
\caption{ 
\label{figure1} 
{\bf Sigmoidal activation. } 
Learning curves for $K=2$ (a,b) and $K=5$ (c,d). 
{\bf (a,c)}: order parameters $R$  and $S$  as functions
of $\alpha=\beta P/(KN)$. 
  {\bf (b,d):}  the  corresponding generalization error $\epsilon_g(\alpha).$
Vertical lines indicate the critical values $\alpha_s(5)\approx 44.3$ (dotted)
and $\alpha_c(5)\approx 46.6$ (dashed), while $\alpha_d(5) \approx 62.8$
is marked by the short vertical line in (d).}
\end{figure}
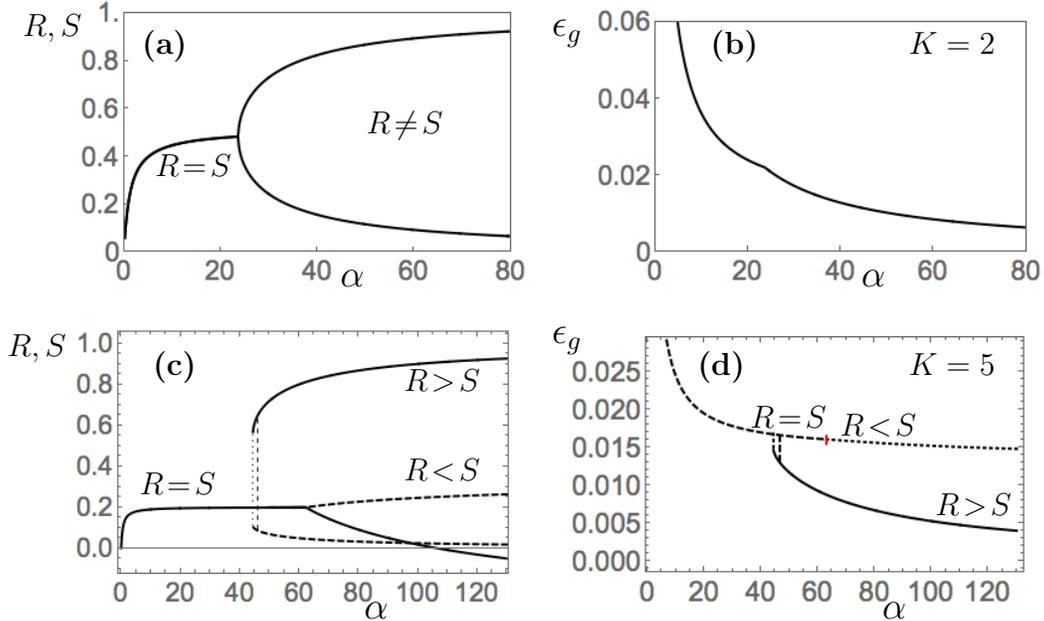

\section{Results and Discussion} \label{results} 
In the following, we present and discuss
our findings for 
the considered student teacher scenarios
and activation functions. 

In order to obtain the equilibrium states of 
the model for given values of $\alpha$ and $K,$ 
we have  minimized the scaled free energy 
(\ref{betafRSC})      
with respect to the site-symmetric order parameters. Potential
(local) minima satisfy the necessary conditions
\begin{equation} \label{zerogradient} 
 \left.{\partial (\beta f)}\right/{\partial R} = 
 \left.{\partial (\beta f)}\right/{\partial C} = 
 \left.{\partial (\beta f)}\right/{\partial S} =  0. 
\end{equation} 
In addition, the corresponding Hesse matrix ${\cal H}$ 
of second derivatives w.r.t.\  $R,S,$ and $C$ has
to be positive definite. This
constitutes  
a sufficient condition for the presence of a local minimum
in the site-symmetric order parameter space. 
Furthermore, we have
confirmed the stability of
the local minima against 
potential 
deviations from  
site-symmetry by inspecting the full matrix
of second derivatives involving the $(K^2\!+\!K(K\!-\!1)/2)$
individual quantities $\left\{R_{ij},Q_{ij}=Q_{ji}\right\}.$

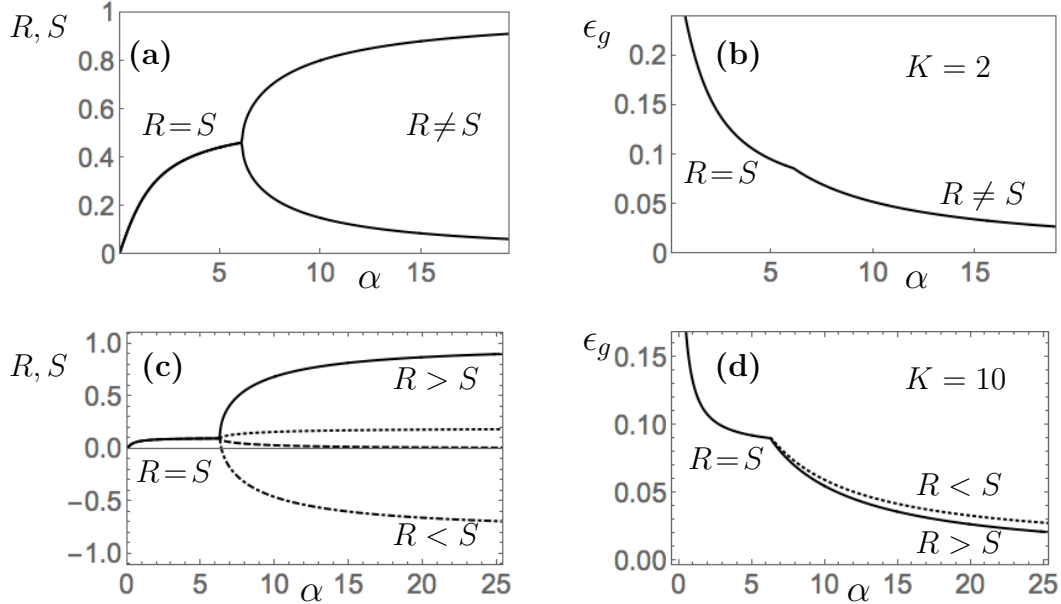
\begin{figure}[t!]
  \begin{center}	
 \hfill
  \includegraphics[width= 0.42\textwidth]{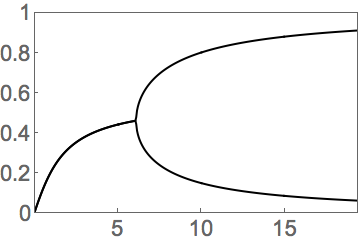} 
 \put(-145,85){{\bf (a)}} 
  \put(-190,95){$R,S$} 
  \put(-40,+58){$R\!\neq\!S$}
  \put(-58,0){\large $\alpha$} 
  \put(-140,58){$R\!=\!S$} 
  \mbox{~~~~~~~~~}
   \includegraphics[width= 0.43\textwidth]{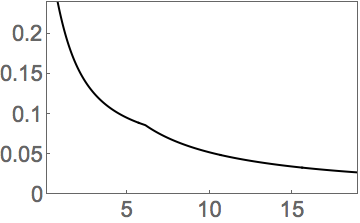}
  \put(-130,85){{\bf (b)}} 
  \put(-58,80){$K=2$}
\put(-45,32){$R\neq S$}
  \put(-142,40){$R\!=\!S$} 
  \put(-180,95){\large $\epsilon_g$} 
  \put(-58,0){\large $\alpha$}  
  \ \\[5mm] 
  \hfill 
  \includegraphics[width= 0.43\textwidth]{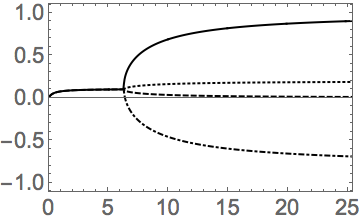}
  \put(-140,85){{\bf (c)}}
  \put(-190,85){$R,S$} 
  \put(-80,0){\large $\alpha$}  
  \put(-45,80){$R>S$}
  \put(-142,45){$R\!=\!S$} 
  \put(-45,21){$R<S$}
  \mbox{~~~~~~~~~}
  \includegraphics[width= 0.43\textwidth]{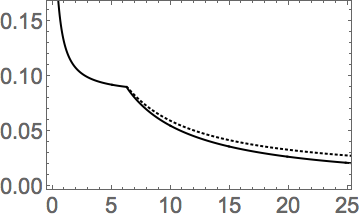} 
 \put(-130,85){{\bf (d)}}
  \put(-58,80){$K=10$}
  \put(-180,95){\large $\epsilon_g$} 
  \put(-140,50){$R\!=\!S$} 
  \put(-80,0){\large $\alpha$} 
  \put(-54,40){$R<S$}
  \put(-54,18){$R>S$}
  \end{center}
\caption{ \label{figure2} {\bf ReLU activation:} 
Learning curves of perfectly matching student   teacher scenarios. 
with $K=2$ (a,b)  and $K=10$ (c,d) 
{\bf (a,c):}  order parameters $R$  and $S$ 
as a function of $\alpha=\beta P/(KN)$. 
{\bf (b,d)}: the corresponding generalization error $\epsilon_g(\alpha).$ 
Specialized solutions with $R>S$
are represented by the solid $(R)$ 
and the dashed line $(S)$ in {\bf (c)}. 
The dotted
line $(S)$ and the chain line $(R)$ represent the local minimum of $\beta f$
with $R<S.$ 
For $K=2$, the transition occurs at $\alpha_c{(2)}\approx 
6.1$, {(a,b)}, while  $\alpha_c(10)\approx 6.2$ (c,d).  
}
\end{figure}


\subsection{Sigmoidal units re-visited} 
The investigation of SCM  with
sigmoidal $g(x)=\mbox{erf}[x/\sqrt{2}]$ 
with $-1 <  g(x) < 1$ 
 along the lines
of the previous section has already
been presented 
in \cite{epl}. A similar model with discrete
binary weights was studied in \cite{Kang}. 

As argued above, for
$g(x)=(1+\mbox{erf}[x/\sqrt{2}])$, the 
mathematical form of the 
generalization error, Eqs.\ (\ref{egsigsiteerf}, \ref{egsigmostgeneral}),
and the free energy $(\beta f)$ are
 the same as for the
 activation $\mbox{erf}[x/\sqrt{2}].$
Hence, the results of 
\cite{epl} carry over without modification. 
The following summarizes
 the key findings of the previous study,
 which we reproduce here for comparison. 

For $K\!=\!2$ we observe that $R=S$ in
thermal equilibrium for small $\alpha$, see 
panels (a) and (b) of  Fig.\ \ref{figure1}.  Both 
hidden units perform essentially the same task and acquire equal overlap with both
teacher vectors, when trained
from relatively small data sets. At a \textit{critical} value
$\alpha_c(2) \approx 23.7$, the system undergoes 
a transition
to a specialized state with $R>S$ or $R<S$ in
which each hidden unit aligns with one specific
teacher unit. 
Both configurations are 
fully equivalent due to the invariance
of the student output under exchange of
the student weights $\mathbf{w}_1$ and
$\mathbf{w}_2$ for $K=2$. 
The specialization process is continuous
with the quantity $|R-S|$ increasing
proportional to
 $(\alpha-\alpha_c(K))^{1/2}$  near the transition. 
This results in a \textit{kink} in the continuous 
learning curve $\epsilon_g(\alpha)$
at $\alpha_c$, as displayed in 
panel (b) of  Figure  \ref{figure1}.

Interestingly, a 
different behavior is found for all $K\geq 3,$ as illustrated for $K=5$
in panels (c) and (d) of Figure \ref{figure1}. 
The following regimes can
be distinguished: 
\begin{itemize}
\item[(a)] $0\leq \alpha < \alpha_s(K)$:~ 
For small $\alpha,$
the only minimum of $\beta f$ corresponds
to unspecialized networks with $R=S$. Within this subspace, a rapid initial decrease of $\epsilon_g$ with $\alpha$ is achieved. 
\item[(b)] $\alpha_s(K)\!\leq\!\alpha\!<\! \alpha_c(K)$:~ 
In $\alpha_s (K)$, a specialized
configuration with $R>S$ appears as
a local minimum of the free energy. 
The $R=S$ 
configuration corresponds to the global minimum
up to $\alpha_c(K).$
At this $K$-dependent  
critical value, the  free
energies of the competing minima coincide. 
\item[(c)] $\alpha\!>\!\alpha_c(K)$:~ Above $\alpha_c$, the 
 configuration with $R>S$ 
constitutes the global minimum
of the free energy and, thus, the 
thermodynamically stable
state of the system. 
Note that the transition from the unspecialized to
the specialized configuration is associated with
a discontinuous change of $\epsilon_g$, cf.\ Fig. \ref{figure1} (d). 
The $(R>S)$ specialized state facilitates 
perfect generalization in the limit
$\alpha\to\infty.$
\item[(d)] $\alpha\!\geq\! \alpha_d(K)$:
In addition, at another characteristic
value $\alpha_d,$ the $(R\!=\!S)$ local minimum
disappears and is replaced by a negatively
specialized  state with  $R<S$. Note that
the existence of this local minimum of
the free energy was not reported in \cite{epl}.
The observed specialization $(S\!-\!R)$ increases
linearly with $(\alpha-\alpha_d)$ for $\alpha\approx\alpha_d$.
This smooth transition
does not yield a \textit{kink} in 
$\epsilon_g(\alpha).$ 
A careful
analysis of the associated Hesse matrix 
shows that the $R<S$ state of poor generalization 
persists for all $\alpha>\alpha_d$, indeed. 
\end{itemize}

The limit $K\to\infty$ with $K\ll N$ has also been 
considered in \cite{epl}: 
The discontinuous transition is found 
to occur  at $\alpha_s(K\to\infty) 
\approx 60.99$ and 
$\alpha_c (K\to \infty) \approx 69.09.$  
Interestingly, the characteristic value
$\alpha_d$ diverges as  $\alpha_d(K)=4\pi K$
for large $K$ \cite{epl}. 
Hence, the additional transition from $R\!=\!S$
to $R\!<\!S$ 
cannot be observed for
data sets of size $P \propto KN$. On this scale,
the unspecialized configuration persists
for $\alpha\to\infty.$
It displays site-symmetric
order parameters 
$R=S={\cal O}(1/K)$  with $R,S>0$
and $C={\cal O}(1/K^2),$ 
see \cite{epl} for details. 
Asymptotically, for $\alpha\to\infty,$ 
they approach
the values $R=S=1/K$ and $C=0$ which
yields the non-zero generalization 
error $\epsilon_g(\alpha\to\infty) =
1/3 - 1/\pi \approx 0.0150.$
On the contrary, the $R>S$ specialized 
configuration achieves $\epsilon_g\to 0,$
i.e.\ perfect generalization, asymptotically.

The presence of a discontinuous 
specialization process
for sigmoidal activations with 
$K\geq 3$ suggests that -- 
in practical training situations --
the network
will very likely be trapped in 
an unfavorable configuration unless 
prior
knowledge about the target is available. 
The escape from the poorly generalizing
metastable state with $R=S$ or $R<S$ requires 
considerable
effort in high-dimensional weight space. 
Therefore, the  success of training will
be  delayed significantly.     
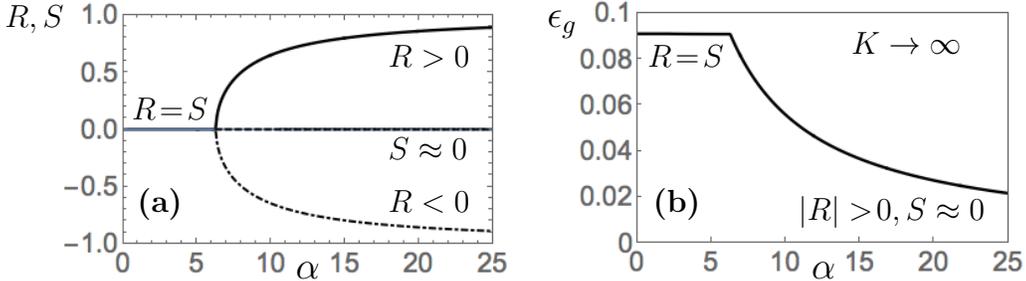
\begin{figure}[t]
  \begin{center}
 \hfill 	
  \includegraphics[width= 0.43\textwidth]{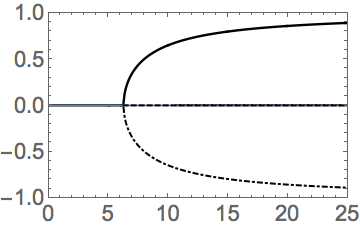} 
 \put(-140,25){{\bf (a)}}
  \put(-190,95){$R,S$} 
  \put(-45,80){$R>0$}
  \put(-142,60){$R\!=\!S$} 
  \put(-45,25){$R<0$}
    \put(-45,45){$S\approx 0$}
  \put(-80,0){\large $\alpha$} 
  \mbox{~~~~~~}
  \includegraphics[width= 0.43\textwidth]{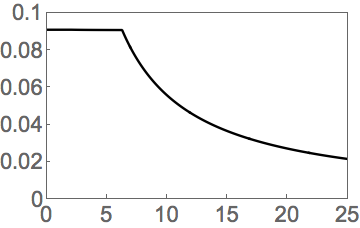}
   \put(-180,95){\large $\epsilon_g$} 
  \put(-142,80){$R\!=\!S$} 
 \put(-140,25){{\bf (b)}}
  \put(-80,0){\large $\alpha$}  
  \put(-65,85){$K\to\infty$}
 \put(-85,22){$|R|>\!0, S\approx 0$} 
 \end{center}
\caption{ \label{figure4} 
{\bf ReLU activation.} 
Learning curves of the perfectly matching student teacher scenario for $K\to\infty$. 
In this limit, the continuous transition occurs at $\alpha_c=2\pi.$  
In panel  {\bf (a)}, the solid line
represents the specialized solution 
with 
$R(\alpha)>0$, while 
the chain line marks the 
solution with $R(\alpha)<0$.
In the former, $S\to 0$ for large
$\alpha$, while 
in the latter, $S$ remains positive 
with  $S={\cal O}(1/K)$
for large $K$. 
The learning curves 
$\epsilon_g(\alpha)$ for the competing 
minima of $\beta f$ coincide
for $K\to\infty$ as 
displayed in {\bf (b)}. }
\end{figure}

\subsection{Rectified linear units} 

In comparison with
the previously studied case 
of sigmoidal activations, we find a 
surprisingly different 
behavior in ReLU networks with $K\geq 3.$

For $K\!=\!2$, our findings parallel
the results for networks
with sigmoidal units:
The network configuration
is characterized by $R\!=\!S$ for $\alpha<\alpha_c(K)$
and specialization increases like  
\begin{equation} \label{criticalexp} 
|R-S|\propto (\alpha-\alpha_c(K))^{1/2}
\end{equation} 
near the transition.  This results in
a \textit{kink} in the learning curve $
\epsilon_g(\alpha)$ at $\alpha=\alpha_c(K)$
as displayed in Fig.\ \ref{figure2} (a,b) for $K=2$ with $ \alpha_c(2)\approx 6.1.$

However, in ReLU networks, the
transition
is also continuous for $K\!\geq \!3$. 
Panels (c,d)   of Fig.\ \ref{figure2}  display the example case
$K=10$ with $\alpha_c({10})\approx 
6.2$.

The student output is invariant under
exchange of the hidden unit weight
vectors, consistent with an
unspecialized state for small $\alpha.$ 
At a critical value $\alpha_c(K)$ the
unspecialized $(R=S)$ configuration 
is replaced by two minima of $\beta f$:
in the global minimum we have $R>S$, 
while the competing local 
minimum corresponds 
to configurations with $R<S.$ 
Only the former facilitates perfect generalization
with $R\to 1, S\to 0$ in the limit
$\alpha\to\infty.$
In both competing
minima the emerging specialization
follows
Eq.\ (\ref{criticalexp}) with 
\textit{critical exponent} $1/2.$

\begin{figure}[t]
  \begin{center}	
\hfill 
  \includegraphics[width= 0.45\textwidth]{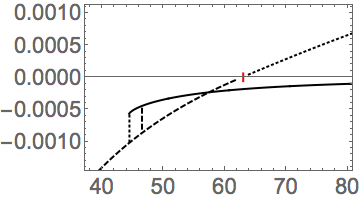} 
 \put(-130,80){{\bf (a)}}
   \put(-185,85){\small $C$} 
  \put(-58,0){\large $\alpha$} 
  \put(-48,80){$R<S$} 
  \put(-96,30){$R\!=\!S$} 
  \put(-40,44){$R>S$} 
  \mbox{~~~~~}
  \includegraphics[width= 0.45\textwidth]{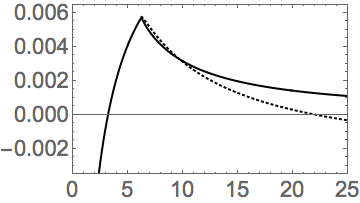} 
 \put(-136,80){{\bf (b)}}
   \put(-185,85){$C$} 
   \put(-46,60){$R>S$} 
    \put(-46,30){$R<S$} 
  \put(-78,0){\large $\alpha$} 
 \end{center}
\caption{ \label{figure3} 
{\bf Student cross overlap} $C$. {\bf (a)} 
 Sigmoidal activations, here $K=5$.
 The values $\alpha_s, \alpha_c, \alpha_d$ 
 are marked as in Fig.\ \ref{figure1} (c).
 For better visibility of the behavior near $\alpha_c$, 
 only a small range of $\alpha$ is shown.
 {\bf (b)} ReLU system with  $K=10$ as an example. 
 }
\end{figure}
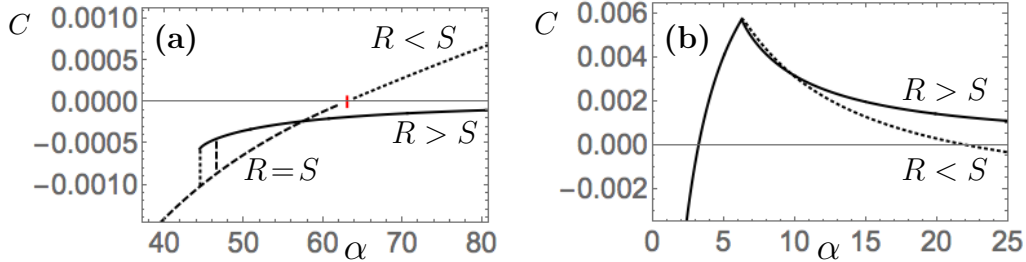

In contrast to the case of
sigmoidal activation, both competing
configurations of the ReLU system
display very similar generalization
behavior. While, in general,  
only states with $R>0$
can perfectly reproduce the teacher
output, the student configurations
with $S>0$ and $R<0$ also achieve 
relatively 
low generalization error for large
$\alpha$, see Fig.\ \ref{figure2}
(c,d) for an example. 

The limiting case of large networks with 
$K\!\to\!\infty$ can be considered
explicitly. 
We find for large ReLU 
networks that the continuous 
specialization 
transition occurs at
$$\alpha_c(K\!\to\!\infty) = 2\pi \approx 6.28.$$ 
The generalization error
decreases very rapidly (instantaneously
on $\alpha$-scale)
from the initial
value of $\epsilon_g (0) \approx 0.341$
with $R\!=\!S\!=\!C\!=\!0$  to a 
plateau with 
$$ \epsilon_g(\alpha)\!=\! \textstyle 
 \frac{1}{4}\!-\!\frac{1}{2\pi} \approx 0.091 
 \mbox{~for~} 0 < \alpha < 2\pi $$
where $R\!=\!S\!=\!1/K \mbox{~~and~~}
C\!=\!{\cal{O}}(1/K^2).$
For $\alpha>\alpha_c$, the 
order parameter
$R$ either increases or decreases
with $\alpha$, approaching the
values $R\to\!\pm 1$ 
asymptotically, while $S(\alpha)=0$
in both branches for $K\to\infty.$

Surprisingly, both solutions display 
the exact same
generalization error, see Fig.\ \ref{figure4} (b). 
Consequently, 
the free energies
$\beta f$ of the competing minima
also  coincide in the limit
$K\to\infty$ since the entropy 
(\ref{entropydef}) satisfies 
$S(-R,0,0)=S(R,0,0).$

In the configuration
with $R<0$ the order
parameters display the scaling 
\begin{equation} \label{antiscaling}
S = {\cal O}\left(1/{K}\right)
\mbox{~~and~~}
C = {\cal O}\left(1/{K^2}\right)
\end{equation}
for large $K.$ 
In \ref{weaknegative}
we show how 
a single teacher ReLU with 
activation $\max(0,x^*)$ can be
approximated by  $(K-1)$ weakly aligned
units in combination with one
anti-correlated student 
node. 
While the former effectively 
approximates a linear response 
of the form $const.\ + x^*$, 
the unit with $R=-1$ implements
$\max(0,-x^*)$. 
Since  
$ \max(0,x^*) = \max(0,-x^*) + x^* $
the student can approximate the 
teacher output very well, see also
the appendix for details.  
In the limit $K\to\infty$, the 
correspondence becomes exact and 
facilitates 
perfect generalization for 
$\alpha\to\infty.$ 

Note that a similar argument does not
hold for student teacher scenarios
with sigmoidal activation functions, 
as they not display the partial
linearity of the ReLU. 

\subsection{Student-student overlaps} 
It is also instructive
to inspect the behavior of
the order parameter $C$ which
quantifies the mutual overlap of 
student weight vectors. 
In the ReLU system with large 
finite $K$,
we observe
$C(\alpha)={\cal O}(1/K^2)>0$ before
the transition. It reaches
a maximum
value at the phase transition 
and decreases
with increasing $\alpha>\alpha_c$. 
In the positively specialized
configuration it approaches the 
limiting value $C(\alpha\to\infty)=0$
from above, 
while it assumes negative values
on the order ${\cal O}(1/K^2)$
in the
configuration with $R<S.$

This is in contrast
to networks of sigmoidal units, where $C<0$
before the discontinuous 
transition and in the specialized $(R>S)$
state, see \cite{epl,epj} for details. 
Interestingly, the characteristic
value $\alpha_d$ coincides with the point
where $C$ becomes positive in the 
suboptimal local minimum of $\beta f.$

Figure \ref{figure3} displays 
$C(\alpha)$ for sigmoidal
(panel a) and ReLU activation (b) for $K=5$ as an example. 
Apparently the ReLU system tends
to favor correlated hidden units
in most of the training process.

\subsection{Monte Carlo simulations} \label{MCS}

In order to demonstrate that our theoretical
results also apply qualitatively in finite systems and beyond the high-temperature limit, 
we performed Monte Carlo simulations of the training processes. 

We have implemented the student teacher scenarios in relatively
small systems with $N=50$ and $K=4$ hidden units. The systems were
trained according to a Metropolis-like scheme with continuous changes
of the student weights. In an individual
Monte Carlo step (MCS), all adaptive weights 
in the student network 
were subject to independent, zero mean additive Gaussian noise
with subsequent normalization to maintain $\mathbf{w}_j^2=N$ for all $j.$  
The associated change $\Delta E$  of the training energy $E$, Eq.\
(\ref{costfunction}), was computed and the randomized modification was
accepted with probability $\min\{1,e^{-\beta \Delta E}\}.$ 
A constant variance of the Gaussian noise was selected as to maintain
acceptance rates in the vicinity of $0.5$ in each setting. 

All simulations were performed with $\beta=1,$ which corresponds to a
relatively low training temperature, and with training set sizes 
$P=\tilde{\alpha}  K N$ that could be handled with moderate computational
effort. 

In principle, all stable and metastable states, i.e.\ local and
global minima of the  associated free 
energy, would eventually be visited starting from arbitrary initializations
by the finite system.  
However, this requires very large equilibration and observation 
times. Therefore we followed an alternative strategy by preparing
initial states which slightly favored  one of the competing configurations
and observed the quasi-stationary behavior of the system at intermediate
training times.  In all training processes, quasistationary states could
be observed after ${\cal O}(10^4)$ elementary MCS. 
Below the specialization transition, the systems approach
unspecialized states independent of the initialization. 
Averages and standard
deviations were determined over the last $1000$ MCS in 20 independent
runs for each  considered setting.

\begin{figure}[t]
 \begin{center}	
  \includegraphics[width= 0.32\textwidth]{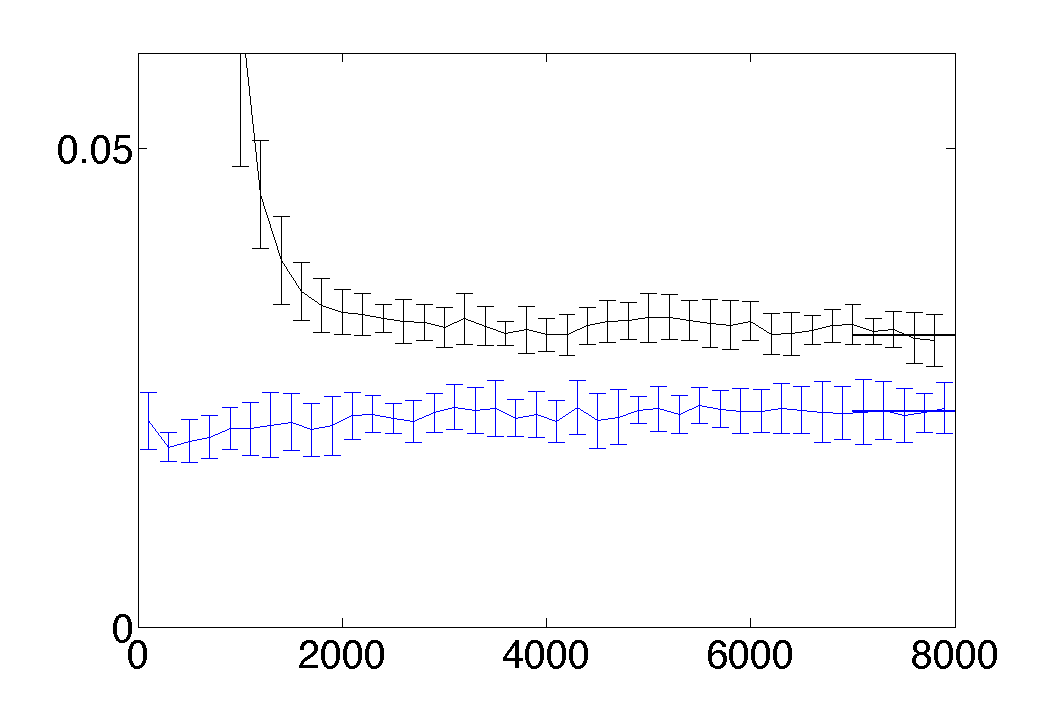} 
\put(-80,60){{\bf (a)}}
  \put(-130,55){\large $\epsilon_g$} 
  \put(-76,-8){\small MCS} 
  \includegraphics[width= 0.32\textwidth]{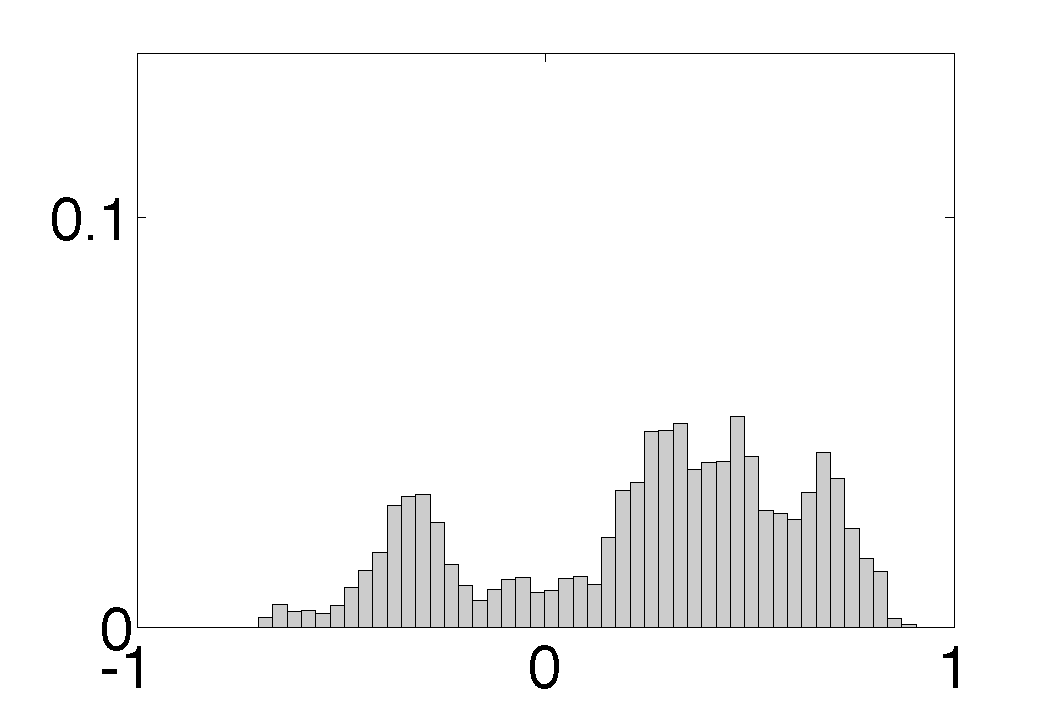}
\put(-95,60){{\bf (b)}}
 \put(-60,-8){$R_{im}$}
  \includegraphics[width= 0.32\textwidth]{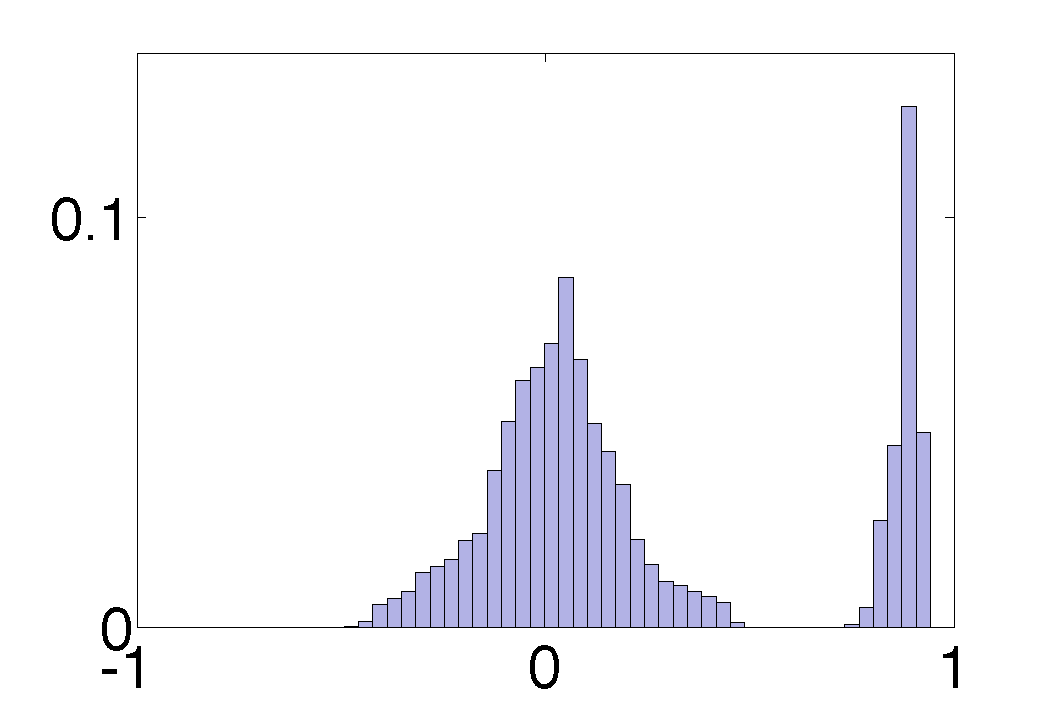}
\put(-95,60){{\bf (c)}}
  \put(-60,-8){$R_{im}$}
 \end{center}
\caption{ \label{figure6} 
{\bf Monte Carlo simulations} of the ReLU system. {\bf (a)} 
the generalization error
as observed with $N=50,\beta=1,K=4$
for $\tilde{\alpha}=24$
 on average over 20 independent runs, error bars represent the
 standard deviations. {\bf (b,c)} histograms of the relative
frequency
of values $R_{im}$ observed over the last 1000 elementary Monte Carlo steps
as marked by the bold solid lines in (a).  
The upper curve in (a) and histogram (b) correspond to initializations of the
systems in slighlty anti-specialized states. For the lower curve and
histogram (c) the systems were initialized with a weak positive specialization,
see Sec. \ref{MCS}. 
}
\end{figure}
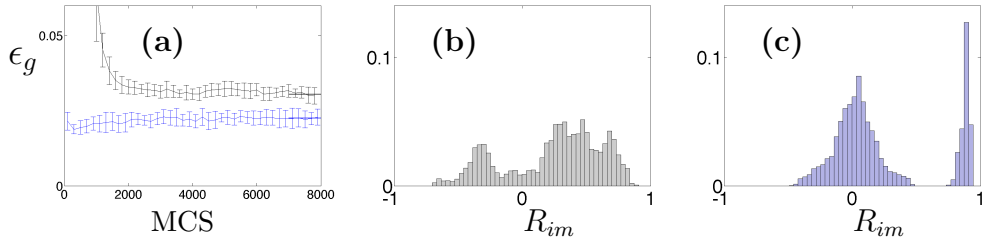

Fig. \ref{figure6} shows example
learning curves in the ReLU system with $K=4$ and training set size 
$\tilde{\alpha}=24,$ which is sufficient for hidden unit specialization. 
The last 1000 MCS
are marked by bold solid lines in panel (a).  The simulations confirm
the existence of two competing quasistationary states. 
Histograms of the observed order parameters $R_{im}$ 
show that they correspond to
a specialized state with few, large positive student teacher overlaps, see
Fig. \ref{figure6} (c). The anti-specialized state is characterized by
a considerable fraction of values $R_{im} <0$, see panel (b). 
We also obtained results for the system with sigmoidal activation, 
which are not displayed here, 
confirming the competition of a specialized state with unspecialized 
configurations. 
Similar findings, including histograms of the observed
$R_{im},$ have been published in \cite{epl} for sigmoidal units only.  
There, the authors also present simulation results for $K=2.$

\begin{figure}[t]
  \begin{center}	
  \includegraphics[width= 0.45\textwidth]{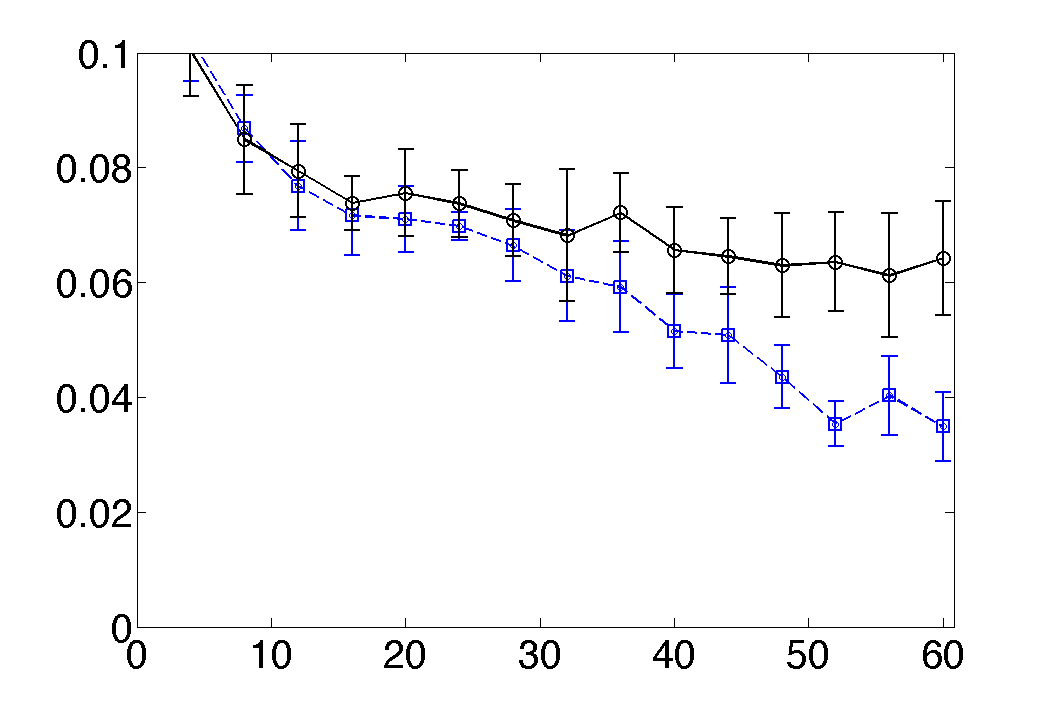} 
\put(-145,25){{\bf (a)}}
  \put(-180,95){\large $\epsilon_g$} 
  \put(-78,-2){\large $\tilde{\alpha}$} 
  \mbox{~~~} 
  \includegraphics[width= 0.45\textwidth]{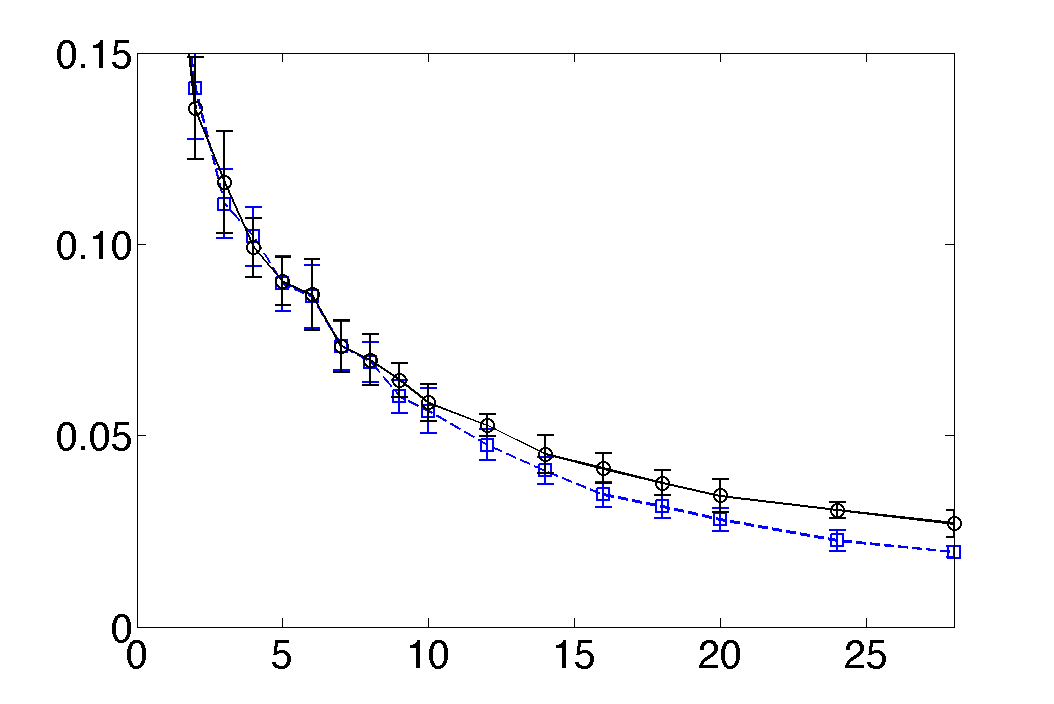} 
\put(-145,25){{\bf (b)}}
  \put(-180,95){\large $\epsilon_g$} 
  \put(-95,-2){\large $\tilde{\alpha}$} 
 \end{center}
\caption{ \label{figure7} 
{\bf Monte Carlo simulations} of the student teacher scenarios.
The generalization error as observed for systems with $N=50,\beta=1,K=4$ 
as a function of $\tilde{\alpha}.$ 
Averages and standard deviations (20 independent simulation runs)
of $\epsilon_g$ were determined
over the last 1000 Monte Carlo steps. 
 {\bf (a)} networks with sigmoidal activation in unspecialized (upper)
 and specialized configurations (lower curve). 
 {\bf (b)} systems with ReLU hidden layer in anti-specialized
 (upper) or specialized (lower) quasistationary states. 
}
\end{figure}
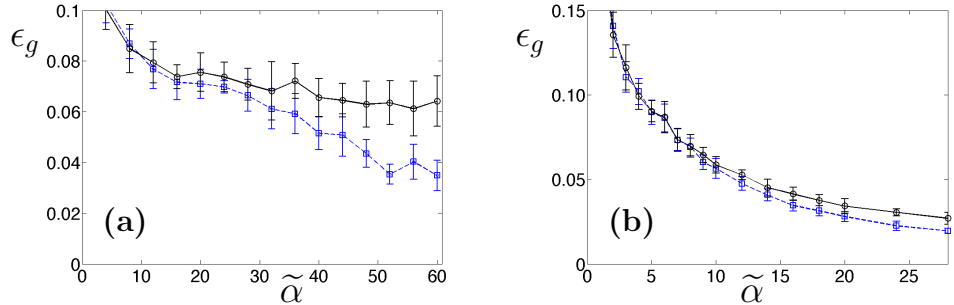

We determined the average 
generalization error from the order parameter values
as observed in the competing quasistationary states of training in 
the last 1000 MCS. Figure \ref{figure7} displays
the corresponding generalization error as a function
of $\tilde{\alpha}$ for sigmoidal activation in panel (a) and for a
ReLU hidden layer in (b). The observed behavior is consistent
with the predicted discontinuous and continuous phase transition, respectively. 
In particular we note that the competing configurations in the 
ReLU system (panel b) display very similar generalization errors. 
In contrast, the difference between specialized and unspecialized sigmoidal 
networks is much more pronounced, see panel (a) of Fig.\ \ref{figure7}. 

While the simulations were performed in fairly small systems and with $\beta=1$,
the results are in very good qualitative agreement with the theoretical predictions obtained in the limits $N\to\infty$ and $\beta\to 0.$
Note that at low temperature, training and
generalization error are not identical. As expected $E/P$ is found to be
systematically lower than $\epsilon_g$. However, we observed that generalization and training error
evolve in parallel with the training time (MCS) and display  
analogous dependencies on the training set size $\tilde{\alpha}$. 
Details will be presented elsewhere.

\subsection{Practical relevance} 

It is important to realize that
a quantitative comparison of the
two scenarios, for instance w.r.t. the 
critical values $\alpha_c$, is not
sensible. 
The complexities of sigmoidal
and ReLU networks with $K$ units 
do not necessarily 
correspond to each other. Moreover, the 
actual $\alpha$-scale is 
trivially related to a potential 
scaling of the activation functions.

However, our results provide valuable
qualitative insight:   
The  continuous nature of the 
transition
suggests that ReLU systems should
display favorable training
behavior in comparison to systems of
sigmoidal units.
In particular, the suboptimal 
competing state displays very good  
performance, comparable to that of the 
properly specialized configuration.
Their generalization
abilities even coincide
in large networks of many hidden units. 

On the contrary, the achievement of 
good generalization in 
networks of sigmoidal units
will  be delayed significantly
due to the  
discontinuous specialization transition
which involves a poorly generalizing 
metastable state.

\section{Conclusion and Outlook}
We have investigated the training of shallow, 
layered neural networks in student teacher scenarios 
of matching complexity. Large, adaptive networks have 
been studied by employing modelling concepts and 
analytical tools borrowed from the statistical 
physics of learning.  Specifically, stochastic 
training processes  at high
formal \textit{temperature} were studied and learning 
curves were obtained for two popular 
types of hidden unit activation. Monte Carlo simulations
confirm our findings qualitatively. 
To the best of our knowledge,  this work constitutes the first theoretical, 
model-based comparison of  sigmoidal activations and 
rectified linear units in feed-forward neural networks.

Our results confirm that networks
with $K\geq 3$ sigmoidal hidden units
undergo a discontinuous 
transition: A critical training set 
size is required
to facilitate the differentiation, i.e. 
specialization 
of hidden units. 
However, a poorly 
performing state of the network persists as a locally
stable configuration for all sizes of the training set. 
The presence of such an unfavorable
local minimum 
will delay successful learning in practice, unless prior 
knowledge of the target rule allows for non-zero initial 
specialization. 

On the contrary, the specialization 
transition is always
continuous in ReLU networks. 
We show that above a 
weakly $K$-dependent 
critical value of the re-scaled
training set size $\alpha $, 
two competing specialized configurations
can be assumed.  
Only one of them displays positive
specialization $R>S$ and
facilitates perfect 
generalization from large training 
sets for finite $K$. 
However, the competing
configuration with negative 
specialization $R<0,S>0$ realizes 
similar performance which is nearly
identical for networks with many 
hidden units and coincides
exactly in the limit  $K\to\infty.$

As a consequence, the problem of
\textit{retarded learning} \cite{retarded} associated 
with the existence of metastable
configurations is expected to
be 
less pronounced in ReLU networks
than in their counterparts with
sigmoidal activation.

Clearly, our approach is subject to 
several limitations which will
be addressed in future studies.
Probably the most straightforward,  
relevant extension of our work
would be the consideration of 
further activation functions,
for instance modifications of the ReLU 
such as
the \textit{leaky} or \textit{noisy} 
ReLU or 
alternatives like \textit{swish} 
and \textit{max-out} 
 \cite{timetoswish,searching}. 

Within the site-symmetric space
of configurations, cf.\ Eq.\ 
(\ref{sitesymmetry}), only the  
specialization of single units
with respect to one of the teacher
units can be considered. 
In large networks, one would expect
partially specialized states, where
subsets of hidden units achieve
different alignment with 
specific teacher units. 
Their study requires the 
extension of the analysis beyond 
the assumption of  site-symmetry.

Training at low formal temperatures 
can be studied
along the lines of \cite{epj} where the 
replica formalism was 
already applied to networks 
with sigmoidal activation. 
Alternatively,
the simpler annealed approximation could
be used \cite{EngelvandenBroeck,SST,revmodphys}. 
Both approaches allow
to vary the control parameter 
$\beta$ of 
the training process and  the scaled 
example set size  
$\widetilde{\alpha}=P/(KN)$  independently,
as it is the case in 
more realistic settings. Note that the findings reported
in \cite{epj} for sigmoidal activation displayed 
excellent qualitative agreement
with the results of the much simpler high-temperature
analysis in \cite{epl}. 

The dynamics of non-equilibrium
on-line training by gradient descent has been studied 
extensively for
soft-committee-machines with sigmoidal activation, e.g.\ 
\cite{saadsollashort,saadsollalong,woehlertransient,nestorplateaus}. 
There, quasi-stationary plateau
states in the learning dynamics
are the counterparts of the 
phase transitions observed 
in thermal equilibrium situations.
First results for ReLU  networks 
have been obtained recently \cite{straat2019}. These
studies should be extended in order to identify and
understand the influence of the activation function on the 
training dynamics in
greater detail.

Model scenarios with mismatched 
student and teacher
complexity will provide further insight into the role of 
the activation function for the learnability of a given
task. It should be interesting to investigate  
specialization transitions in 
practically relevant settings
in which either the task is unlearnable
$(K<M)$ or
the student architecture is over-sophisticated 
for the problem at hand ($K>M$). In addition,
student and teacher systems with
mismatched activation functions should 
constitute interesting model systems.

The complexity of the 
considered networks can be increased
in various directions. If the simple shallow architecture
of Eq.\ (\ref{studentoutput}) is extended by
local thresholds and hidden to output weights
that are both adaptive, 
it parameterizes a \textit{universal approximator}, 
see e.g.\ 
\cite{cybenko,hornik,hanin}. Decoupling the 
selection of these few additional parameters from the 
training of the input to hidden weights 
should be possible following
the ideas presented in \cite{adiabatic}. 

Ultimately,  \textit{deep}
layered architectures
should be investigated along the same lines. As a 
starting point,
simplifying tree-like architectures  could be considered as 
in e.g.\ \cite{retarded,relucapacity}.

Our modelling approach and 
theoretical analysis goes beyond the 
empirical investigation of data set 
specific performance. The suggested
extensions bear the promise to
contribute to a better, fundamental 
understanding of layered neural networks
and their training behavior. 

\section*{Acknowledgements}
M.S. and M.B. acknowledge   financial support through the Northern Netherlands Region of
Smart Factories (RoSF) consortium, led by the Noordelijke Ontwikkelings en Investerings Maatschappij
(NOM), see { \tt{www.rosf.nl}}. 

\appendix  \label{appendix}

\section{Mathematical Details}
\subsection{Co-variance matrix and order parameters}
\noindent
The $(K\!+\!M)\!\times\!(K\!+\!M)$-dim.\ 
matrix of order parameters reads
\begin{equation} \label{fullmatrix}
    {\cal C} = 
    \left[ \begin{array}{cc} {T} & {R}\\ R^\top & Q
    \end{array}  \right]
    \mbox{~with submatrices~} 
 T\! \in\! \mathbb{R}^{M\times M}, 
 R\!\in\! \mathbb{R}^{K\times M}, 
 Q\!\in\! \mathbb{R}^{K\times K} 
\end{equation} 
of elements 
$T_{ij}\!=\!\left\langle x^*_i x^*_j\right\rangle\! =\!\frac{\mathbf{w}_i^*\cdot
\mathbf{w}_j^*}{N}$, 
$R_{ij}\!\!=\!\left\langle x_i x^*_j\right\rangle\!=\!\frac{\mathbf{w}_i\cdot
\mathbf{w}_j^*}{N},$ and
$ Q_{ij}=\!\left\langle x_i x_j\right\rangle\!=\! \frac{\mathbf{w}_i\cdot
\mathbf{w}_j}{N}.$
Note that Eqs.\ (\ref{slogdet}) and (\ref{sK}) 
correspond to the special case of $K=M$ and exploit site-symmetry (\ref{sitesymmetry})
and normalization (\ref{normalization}).

\subsection{Derivation 
of the generalization error}
Here we give a derivation of the generalization error in terms of the order parameters for sigmoidal and ReLU student and teacher. 
For general $K$ and $M$ it reads
\begin{equation} \label{epsggeneral_v2}
    \epsilon_g = \frac{1}{2K}
    \Bigg(\sum_{i,j=1}^K \langle g(x_i) g(x_j) \rangle -2 \sum_{i=1}^K \sum_{j=1}^M \langle g(x_i) g(x_j^*) \rangle \,\, + \sum_{i,j=1}^M \langle g(x_i^*) g(x_j^*) \rangle \Bigg), 
\end{equation}
which reduces to Eq.\ (\ref{epsggeneral}) for $K=M.$ 
To obtain $\epsilon_g$
for a  particular choice of activation function $g$, expectation values of the form $\left\langle g(x)g(y) \right\rangle$ have to be evaluated over the joint normal 
density of the hidden unit local potentials $x$ and $y$, i.e.\  $P(x,y)=\mathcal{N}(\mathbf{0},\widehat{\mathcal{C}})$ with the
appropriate submatrix
$\widehat{\mathcal{C}}$ of 
${\cal C}$, cf. Eq.\ (\ref{fullmatrix}): 
$$ \widehat{\mathcal{C}}=
\left[
\begin{array}{cc}	
\left\langle y^2\right\rangle &
\left\langle x y\right\rangle\\
\left\langle x y \right\rangle 
 & \left\langle x^2 \right\rangle
\end{array}
\right].
$$
\subsubsection{Sigmoidal}
For student and teacher with sigmoidal activation functions $g(x)=\mbox{erf}\big[\frac{x}{\sqrt{2}}\big]$ or $g(x)=\left(1+\mbox{erf}\big[\frac{x}{\sqrt{2}}\big]\right)$, the generalization error has been derived in \cite{saadsollalong}: 
\begin{eqnarray} \label{egsigmostgeneral}  
\epsilon_g  &=& 
\frac{1}{\pi} \left\{
\sum_{i,j=1}^K \sin^{-1}\frac{Q_{ij}}{\sqrt{1+Q_{ii}}\sqrt{1+Q_{jj}}}
\right.
+ 
\sum_{n,m=1}^M \sin^{-1}\frac{T_{nm}}{\sqrt{1+T_{nn}}\sqrt{1+T_{mm}}}  
\nonumber \\
&& - \textstyle  
\left. 2 \sum_{i=1}^K \sum_{j=1}^M \sin^{-1}\frac{R_{ij}}{\sqrt{1+Q_{ii}}\sqrt{1+T_{jj}}}
 \right\}.
\end{eqnarray} 
\subsubsection{ReLU} 
For student and teacher with ReLU activations $ g(x)= \mbox{max}\{0,x\}$, applying the elegant formulation  used in \cite{yoshida_statistical_2017} gives an analytic expression for the two-dimensional integrals:
 $  \langle g(x) g(y) \rangle
    = \left \langle \mbox{max}\{0,x\} \mbox{max}\{0,y\} \right \rangle$
\begin{equation} \label{expvaltermsrelu} \textstyle
      = \int\limits_0^\infty \int\limits_0^\infty x y \mathcal{N}(\mathbf{0},
    \widehat{\mathcal{C}}) dx dy 
    = \frac{\widehat{\mathcal{C}}_{12}}{4} + 
\left. {\sqrt{
    \widehat{\mathcal{C}}_{11} \widehat{\mathcal{C}}_{22} - \widehat{\mathcal{C}}_{12}^2} + \widehat{\mathcal{C}}_{12} \sin^{-1}\left[ \frac{\widehat{\mathcal{C}}_{12}}{\sqrt{\widehat{\mathcal{C}}_{11}\widehat{\mathcal{C}}_{22}}} \right]}\right/{2\pi}.
\end{equation}
Substituting the result from Eq. (\ref{expvaltermsrelu}) in Eq. (\ref{epsggeneral_v2}) for the corresponding covariance matrices gives the analytic expression for the generalization error in terms of the order parameters:
\begin{eqnarray} \label{egrelumostgeneral}
 \epsilon_g &=&  \frac{1}{2K} \sum_{i,j=1}^K \left( 
 \frac{Q_{ij}}{4} \!+\! \frac{\sqrt{Q_{ii}Q_{jj}\!-\!Q_{ij}^2}\!+\!Q_{ij} \sin^{-1}\left[\frac{Q_{ij}}{\sqrt{Q_{ii}{Q_{jj}}}}\right]}{2\pi}
  \right)  
 \\  && 
 - \frac{1}{K} \sum_{i=1}^K \sum_{j=1}^M \textstyle \left( \frac{R_{ij}}{4}\!+\! 
 \frac{\sqrt{Q_{ii}T_{jj}\!-\!R_{ij}^2}\! +  \!R_{ij}\sin^{-1}\left[\frac{R_{ij}}{\sqrt{Q_{ii}T_{mm}}}\right]}{2\pi}
 \right)
 \nonumber \\
 && 
+ \frac{1}{2K} \sum_{i,j=1}^M \textstyle \left( \frac{T_{ij}}{4} \!+\!  \frac{\sqrt{T_{ii}T_{jj}\!-\!T_{ij}^2}\!+\!T_{ij} \sin^{-1}\left[\frac{T_{ij}}{\sqrt{T_{ii}T_{jj}}}\right]}{2\pi} \right).
\end{eqnarray} 
For $K=M$, orthonormal teacher vectors with $T_{ij}=\delta_{ij}$, fixed student norms $Q_{ii}=1$, and assuming site symmetry, Eq.\ 
(\ref{sitesymmetry}), 
we obtain  Eqs. (\ref{egsigsiteerf}) and (\ref{egsigsiterelu}), respectively.

\subsection{Single unit student
and teacher} 
In the simple case $K=1$ with a single unit as student
and teacher network, we have to consider only one order 
parameter $R=\mathbf{w}\cdot\mathbf{w}^*/N.$ Assuming $\mathbf{w}\cdot\mathbf{w}= \mathbf{w}^*\cdot\mathbf{w}^*=N,$
we obtain the free 
energy $\beta f=\alpha\, \epsilon_g -s$ with
\begin{eqnarray} 
s&=&  \frac{1}{2} \ln [1-R^2] \, + \, const.\ \hfill  \ \\
\epsilon_g &=& \frac{1}{3}- \frac{2}{\pi} \sin^{-1}\left[{R}\big/{2}\right] 
\mbox{~~~~~~~~~~~~~(sigmoidal)~~~}\\
\epsilon_g &=& \frac{2\!-\!R}{4}\!-\! \frac{\sqrt{1\!-\!R^2}\!+\!R \sin^{-1}[R]}{2\pi}
\hfill \mbox{~~~~~(ReLU).~~}
\end{eqnarray} 
The necessary condition
$\partial{(\beta f)}/{\partial R}
{=}0$ becomes
\begin{eqnarray} 
\alpha &=&  \frac{\pi R \sqrt{4-R^2}}{2 (1-R^2)} 
\hfill \mbox{~~~~~~~~~~~~~~~~~~(sigmoidal)~}\\[1mm]
\alpha &=& 
\frac{4\pi R}{(1-R^2) (\pi +
2 \sin^{-1}[R])}
\hfill \mbox{~~~~~~~(ReLU).}
\end{eqnarray} 
In both cases, the student teacher overlap increases
smoothly from
\textit{zero} to $R=1.$ 
A Taylor expansion
of $1/\alpha$ for $R \approx 1$ yields the
asymptotic behavior
$$ R(\alpha)= 1- \frac{const.}{\alpha }
\mbox{~~and~~} 
\epsilon_g(\alpha) = 
\frac{1}{2\alpha} \mbox{~~for
~} \alpha\to \infty$$
for both types of activation. 
This basic large-$\alpha$
behavior with
$\epsilon_g \propto \alpha^{-1}$ 
carries over to the specialized solutions for settings with $K\!=\!M.$ 

\subsection{Weak and negative
alignment} \label{weaknegative} 

Here we consider a particular 
teacher unit which realizes a 
ReLU response
$\max(0,x^*) \mbox{~with~}
x^*=\mathbf{w}^*\cdot 
\boldsymbol{\xi}/\sqrt{N}.$
A set of $K$ hidden units in the student
network can obviously reproduce the
response by aligning one of the units
perfectly with, e.g., 
 $ 
R\!=\!\mathbf{w}_1\cdot\mathbf{w}^*/N\!=\!1$ and 
$S\!=\!\mathbf{w}_j
\cdot\mathbf{w}^*/N\!=\!0$  for $j>1$.
Similarly, we obtain for $R=-1$ that
$
x_1 = -\mathbf{w}^* \cdot \boldsymbol{\xi} 
\mbox{~~and~} \max(0,x_1)=\max(0,-x^*).
 $

Now consider 
the mean response of a
student unit with small positive
overlap 
$S=\mathbf{w}_j\cdot\mathbf{w}^*/N$,
given the teacher unit response $x^*$.
It corresponds to
the average $\left\langle g(x_j) \right\rangle_{x^*}$ over  the conditional density
$ P(x_j|x^*) =
P(x_j,x^*)/P(x^*)$. 
One obtains
$$ \left\langle g(x_j)
\right\rangle_{x^*} =\, {1}/{\sqrt{2\pi}} +{S}
\, x^* / 2  + {\cal O}(S^2)$$ 
by means of a Taylor expansion for 
$S\approx 0$.
As a special case, 
the  mean  
response of an orthogonal
unit with $S=0$ is $1/\sqrt{2\pi},$
independent of $x^*.$

It is straightforward to work out the 
conditional average of the total 
student response 
for a particular order parameter
configuration with 
$R=-1$ and $S={2}/{(K-1)}.$
Apart from the prefactor $1/\sqrt{K}$
it is given by
$$ 
\max(-x^*,0)  + x^* + \textstyle
\frac{K-1}{\sqrt{2\pi}}
  =  \max(0,x^*) + 
  \frac{K-1}{\sqrt{2\pi}},
$$
where the right hand side coincides
with the expected 
output for $R=1$ and $S=0.$
Hence, the average response agrees
with the teacher output for large $K$. 
Moreover, the correspondence becomes 
exact 
in the limit $K\to\infty,$ which
facilitates perfect generalization
in the negatively specialized state 
with $S>0,R<0$ discussed  in Section
\ref{results}.




\section*{References} 
\bibliographystyle{elsarticle-num} 
\biboptions{sort&compress} 
\bibliography{biehlrelu.bib}


\end{document}